%% file: main.tex
\documentclass{article}

\usepackage[final, nonatbib]{nips_2016}

\usepackage[utf8]{inputenc} 
\usepackage[T1]{fontenc}    
\usepackage{hyperref}       
\usepackage{url}            
\usepackage{booktabs}       
\usepackage{amsfonts}       
\usepackage{nicefrac}       
\usepackage{microtype}      

\usepackage{graphicx}
\usepackage{amsmath}
\usepackage{amssymb}
\usepackage{mathptmx}
\usepackage{graphicx}
\usepackage{subfigure}
\usepackage[textwidth=2cm,colorinlistoftodos]{todonotes}
\usepackage{xspace}
\usepackage{latexsym}
\usepackage{amsmath,amssymb,graphicx,xspace}
\usepackage{times}   
\usepackage{multirow}
\usepackage{footnote}
\usepackage{lipsum}

\usepackage{color,comment,rotating,subfigure,url,xspace} 
\usepackage{tikz}
\usetikzlibrary{arrows,shadows,shapes,backgrounds,decorations,snakes,fit}
\usepackage{todonotes}
\usepackage{enumerate}

\newcommand\blfootnote[1]{%
  \begingroup
  \renewcommand\thefootnote{}\footnote{#1}%
  \addtocounter{footnote}{-1}%
  \endgroup
}
\title{ Diagnostic Prediction Using Discomfort Drawings}

\author{
Cheng Zhang$^*$~\qquad~Hedvig Kjellstr\"{o}m$^*$~\qquad~Bo C. Bertilson$^{\dag}$%
  \\
    {\texorpdfstring{%
      \footnotesize
      \begin{minipage}[t]{.4\textwidth}
	  \vspace{0.1cm}
	  \centering
	  $^*$Robotics, Perception and Learning (RPL)\\
	  KTH Royal Institute of Technology\\
	  Stockholm, Sweden\\
	  \tt\small{\{chengz, hedvig\}@kth.se} \\
      \end{minipage}%
      \begin{minipage}[t]{.65\textwidth}
	  \vspace{0.1cm}
	  \centering
	  $^{\dag}$Department of Neurobiology, Care Sciences and Society\\
	  Karolinska Institute \\
	  Stockholm, Sweden\\
	  \tt\small{bo.bertilson@karolinska.se}\\
      \end{minipage}}{The Author}
   }
}

\begin{document}

\maketitle
\vspace{-1.4cm}
\blfootnote{\scriptsize This work has recently been published in Machine Learning in Health Care conference, 2016. This paper extends the model of the published work to a three-view model with new experimental results. This research has been supported by the Swedish Research Council (VR) and  Stiftelsen Promobilia.   }

\begin{abstract}
\vspace{-5pt}
In this paper, we explore the possibility to apply machine learning to make diagnostic predictions using discomfort drawings. A discomfort drawing is an intuitive way for patients to express discomfort and pain related symptoms. These drawings have proven to be an effective method to collect patient data and make diagnostic decisions in real-life practice. 
A dataset from real-world patient cases is collected for which medical experts provide diagnostic labels. 
Next, we extend a factorized multimodal topic model, Inter-Battery Topic Model (IBTM),  to train a system that can make diagnostic predictions given an unseen discomfort drawing.  Experimental results show reasonable predictions of diagnostic labels given an unseen discomfort drawing. 
The positive result indicates a significant potential of machine learning to be used for parts of the pain diagnostic process and to be a decision support system for physicians and other health care personnel.
\end{abstract}

\vspace{-0.3cm}
\input{intro.tex}
\input{probstate.tex}

\input{model.tex}

\input{exp.tex}
\input{discussion.tex}

%

\small
\bibliography{ref}
\bibliographystyle{plain}

\end{document}

%% file: intro.tex
\vspace{-8pt}
\section{Introduction}
\label{sec:intro}
\vspace{-8pt}

A discomfort drawing is a drawing on the image of a body where a patient may shade all areas of discomfort in preparation for a medical appointment. The drawing has been shown to be able to make diagnostic predictions - especially to discern neuropathic from nociceptive and psychiatric diseases  \cite{bertilson2007pain}. The use of drawings (pain drawing) to collect data from patients was first reported by Palmer in 1949 \cite{palmer1949pain} and has been studied in clinical settings showing high diagnostic predictive value especially in spine related pain by  \cite{ohnmeiss1999relation, vucetic1995pain, albeck1996critical, tanaka2006cervical}. The pain drawing, where different signs mark different kind of pain, is still in use at many clinics. As a more recent method, the discomfort drawing (a revised pain drawing) instructs the patient to shade all areas of discomfort. This method  may have some possible benefits  due to the fact that many different symptoms may arise from disfunction of  the same body organ or nerve \cite{bertilson2003reliability,bertilson2007pain,bertilson2010assessment}.  
Hence, we focus on the use of  discomfort drawings. 

To find high-quality diagnostic prediction methods is a goal of health care  as well as the machine learning community. 
However, the most common medical problem, unspecific pain and discomfort \cite{upshur2010they}, to which machine learning has not been applied yet. In this paper, we focus on applying machine learning for diagnosing pain-related problems using discomfort drawings. 

Topic models \cite{blei03latent}, a type of generative models, have been successfully applied in different domains.
With efficient inference algorithms \cite{hoffman2010online, ranganath2013adaptive},  these models can handle both  small and big datasets, in complete data and in incomplete scenarios.    Additionally, they are highly interpretable and can be used to generate missing data.
In our application of using discomfort drawings for diagnostic prediction, the data consist of multiple modalities (drawings and different types of labels). 
 A recent advancement in multi-modal topic models shows that Inter-Battery Topic Model (IBTM)~\cite{zhang16IBTM} is robust to noise in the data by explaining away irrelevant parts of the information.
Therefore, in this paper IBTM is adapted to predict diagnostic labels given a discomfort drawing. To enhance the interpretability of different type of diagnostic labels, we extend the original two view IBTM to three views.  IBTM was originally proposed for representation learning and applied for classification tasks. In this paper, we adapt the framework for diagnostic label prediction and use mean-shift clustering \cite{comaniciu2002mean} to determine the number of diagnostic predictions that the system needs to make.

The main contribution of this paper lies in the modification and use of  generalized IBTM for diagnostic prediction with discomfort drawings. 
This is a novel application of a principled framework. For this purpose, a dataset was collected from real-world clinical cases with medical expert labels. The experiments show that the adapted IBTM makes reasonable diagnostic predictions.  Additionally, the model also contributes to the interpretability of the data for humans and may further provide insight into the diagnostic procedure. Our approach shows that the use of machine learning in the assessment of discomfort drawings is a promising direction.

%% file: probstate.tex
\vspace{-3pt}
\section{Problem Statement}
\vspace{-6pt}
\begin{table}[h]
\centering
\begin{tabular}{  c  p{9cm} }
\multirow{3}{*}{
\includegraphics[height=4cm]{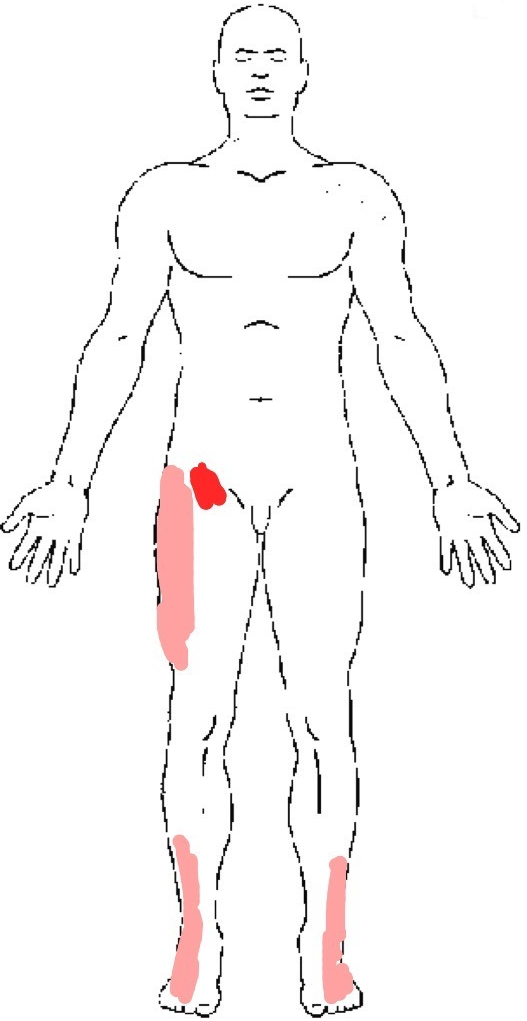}
\includegraphics[height=4cm]{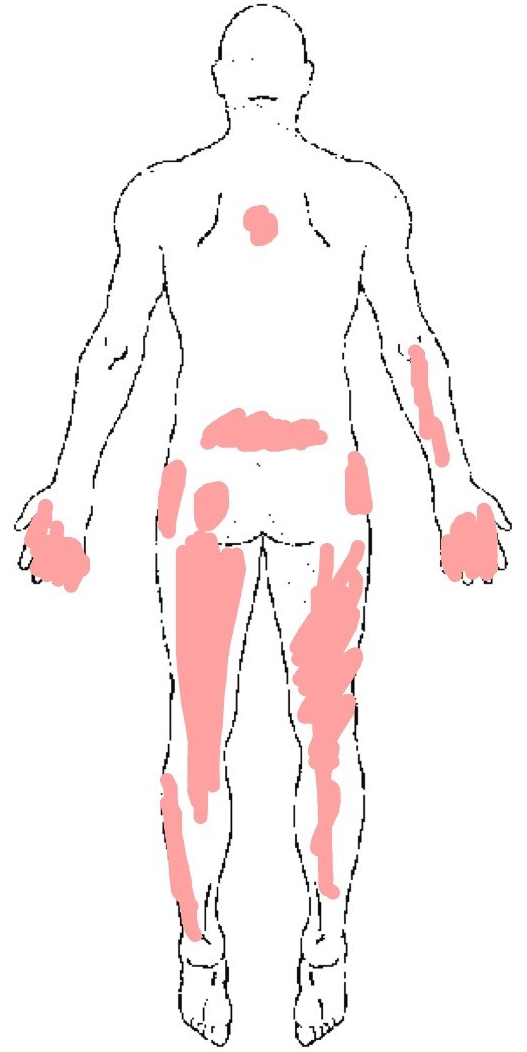}
}
&\textbf{Symptom diagnoses:}
Interscapular discomfort; R arm discomfort; B hands discomfort; Lumbago; B crest of the ilium discomfort; L side thigh discomfort; B back thigh discomfort; B calf discomfort; B achilles tendinitis; B shin discomfort; R inguinal discomfort;\\[5pt]
& \textbf{Pattern diagnoses}
B L5 Radiculopathy; B S1 Radiculopathy; B C7 Radiculopathy;\\[5pt]
& \textbf{Pathophysiological diagnoses}
DLI L4-L5; DLI S1-S2; DLI C6-C7\\[5pt]
\end{tabular}
\caption{\footnotesize Discomfort drawings (left) and diagnoses by medical expert (right). R stands for right-side, L stands for left-side and B stands for bilateral.  DLI refers to discoligament injury.}
\label{fig:dataExp}
\end{table}
\vspace{-12pt}

At some clinics, 
a patient is asked to shade all areas of discomfort on a drawing of a body. 
The patient is typically also asked to specify what type of discomfort they experience and furthermore to describe the discomfort-level over time. During a patient interview additional  information regarding symptoms, prior treatment and  experiences may be added to provide the health care personnel with sufficient information to make a diagnostic prediction that can guide the treatment.

In this paper we focus on diagnostic prediction solely based  on areas of discomfort which is the key information.
Table \ref{fig:dataExp} shows an example of discomfort drawings and their diagnoses.  
On a standard body contour the discomfort regions are marked in red. 
The right column shows the diagnostic label provided by medical experts which are roughly ordered by symptom diagnoses, possible pattern diagnoses and possible pathophysiological diagnoses.
Our task is to build a system that makes high quality diagnostic predictions given a discomfort drawing. 
This could be extended into a  decision support system, which could increase the effectiveness and precision of the care for a large group of less favored patients \cite{upshur2010they}.

%% file: model.tex
\vspace{-5pt}
\section{Model}
\vspace{-5pt}
For this application, we adapt IBTM, which is a generative model.
One advantage of  generative models is that they achieve good performance even on small data sets. 
As it is expensive to collect data in the health care system and there is a big variance in the frequency of different types of diseases, this is highly important. Secondly, generative models have the advantage of being able to handle missing data. 
In this preliminary work, we are only dealing with two modalities, discomfort drawings and diagnostic labels. 
Even in this simplistic setting, the diagnostic labels are not complete. 
In health care systems, there exists a variety of  examinations and tests that are only partially used for different patients.
Hence, a system that can handle missing data is desired in such application. 
Finally, a probabilistic interpretation of the symptoms and diagnostic decisions is desirable. 
IBTM is a factorized multi-modal topic model which enjoys all the properties of generative models and is robust to noise in the data.
  
  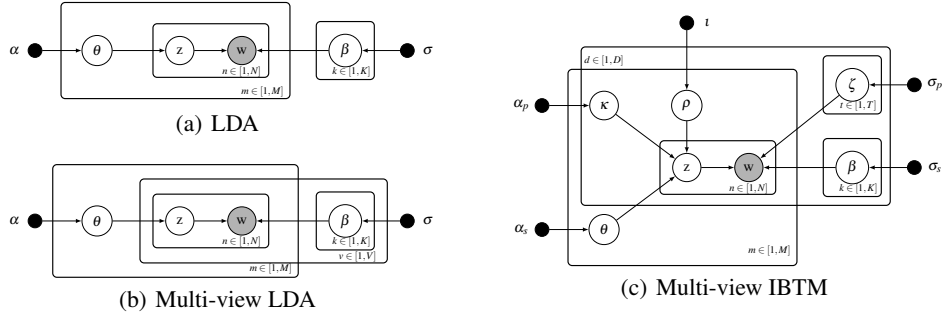
\begin{figure}[h]
  \begin{minipage}{.47\textwidth}
\centering
\subfigure[LDA]{
\scalebox{0.55}{\input{tikz/ldaPGM.tex}}
}
\subfigure[Multi-view LDA]{
\scalebox{0.55}{\input{tikz/PGM_NView_LDA}}
}
\end{minipage}~~~
\begin{minipage}{.47\textwidth}
\subfigure[Multi-view  IBTM]{
\scalebox{0.55}{\input{tikz/general_PGM.tex}}
}
\end{minipage}%
\vspace{-10pt}
\caption{\footnotesize Graphical representations of three topic models.The nodes with grey shadows indicate an observation while all other nodes are latent variables that need to be learned.}
\label{fig:FMLDA}
\end{figure}

\vspace{-5pt}
\paragraph{ Inter-Battery Topic Model} 
\vspace{-5pt}

 IBTM is based on LDA which is shown in Figure~\ref{fig:FMLDA}(a). 
LDA assumes that each word in a document is generated by sampling from a per document topic distribution $\theta \sim Dir(\alpha)$ and per topic word distribution $\beta \sim Dir(\sigma)$. 
LDA is designed for data with a single modality. 
In our application, we want to  jointly model discomfort drawings and the diagnostic labels where each patience case corresponds to one document. 
Additionally, we want to learn a system that is able to give high quality predictions of diagnostic labels given an unseen discomfort drawing. Hence, a multi-modal topic model is needed. Multi-view LDA {\cite{blei2003modeling} as shown in Figure~\ref{fig:FMLDA}(b) is a  extension of LDA to multi-view scenario, where plate $V$ shows different views and $\theta$ is the joint latent representation.


As shown in Figure \ref{fig:FMLDA}(b),  Multi-view LDA forces the different modalities to completely share a latent space $\theta$. However,  real-life data is noisy and incomplete in general and might have shared and disjunct latent sources. 
 IBTM is proposed to be more robust with respect to complex real-life data. Compared to Multi-view LDA, a private topic space for each modality ($\kappa$)  is introduced to explain away irrelevant information. By this, the  shared topic space can provide qualitatively better latent representations of the structure of the data.  In IBTM, $\rho \sim Beta(\iota_{1})$  models portions of the information that can be shared between different modalities, where $\iota_{1}$ and $\iota_{2}$ are two dimensional pairs of beta distribution hyper-parameters.

In our task, we uses three view IBTM. The first view is the discomfort drawing.
For the diagnostic labels, we can see that symptoms diagnoses is just medical description of the symptoms, while pattern diagnoses and pathophysiological diagnoses analysis the reason of these symptoms. Hence, we want to differ these two type of diagnostic labels. Additionally, given an symptom label, we would like to know the reason causing the symptom.  
Therefore, the second view is the symptom diagnostic labels  and the third view is the reason diagnostic labels. Both views  are only available in the training phase. 
Each document $m$ contains a discomfort drawing and its corresponding symptom diagnostic labels and reason diagnostic labels. 
All three modalities share the same per document topic distribution $\theta$ which can be interpreted as the combination of symptoms that generate a drawing and its diagnostic labels. 
For each modality, the private topic distributions $\kappa$ are used to encode the information that cannot be simultaneously explained by all modalities which are noises per se  in general.

To learn all  latent parameters in IBTM, mean field variational inference is used in this work, because variational inference is efficient and can easily be adapted to online settings \cite{hoffman2010online,ranganath2013adaptive,wang2011online}. 
In real health applications, online learning is desirable. 
\vspace{-5pt}
\paragraph{Diagnostic Prediction using IBTM} 
\label{sec:DPIBTM}
\vspace{-5pt}
In the training phase, all latent variables will be learned. 
In the testing phase, given an unseen observation $w$ without diagnostic labels, we will estimate the per document distributions $\theta$ 
 and then generate possible diagnostic labels for the other views. 
Using IBTM, we can generate all possible diagnostic labels for a drawing with different probabilities. However, it is difficult to decide how many diagnostic labels are actually  needed since there exists no universal probability threshold.
We thus use the mean shift clustering \cite{comaniciu2002mean} to cluster the marked area and use the number of clusters as the number of output for each view. 

Besides automatic diagnostic prediction,  it is useful to investigate which features the models learn to make these diagnostic predictions. Based on big datasets, the model may give us insights into diagnostic procedures. Hence,  we can also  generate synthetic discomfort drawings given a diagnostic label and generate the reason diagnoses based on and symptom label. 

%% file: tikz/ldaPGM.tex
\pgfdeclarelayer{background}
\pgfdeclarelayer{foreground}
\pgfsetlayers{background,main,foreground}

\begin{tikzpicture}

\tikzstyle{surround} = [thick,draw=black,rounded corners=1mm]

\tikzstyle{scalarnode} = [circle, draw, fill=white!11,  
    text width=1.2em, text badly centered, inner sep=2.5pt]

\tikzstyle{scalarnodeCyan} = [circle, draw=cyan, fill=white!11,  
    text width=1.2em, text badly centered, inner sep=2.5pt]
\tikzstyle{discnode}=[rectangle,draw,fill=white!11,minimum size=0.9cm]

\tikzstyle{Vnode}=[circle, radius=0.2 cm,draw,fill=black]
\tikzstyle{vectornode} = [circle, draw, fill=white!11,  
    text width=2.3em, text badly centered, inner sep=2pt]
\tikzstyle{state} = [rectangle, draw, text centered, fill=white, 
    text width=8em, text height=6.7em, rounded corners]

\tikzstyle{arrowline} = [draw,color=black, -latex]
\tikzstyle{carrowline} = [line width=2pt, draw,color=black, -latex]
\tikzstyle{line} = [draw]

\node [Vnode] at ( 0.5, 1.5) (alpha_s){};
\node [] at ( 0, 1.5) (){$\alpha$};
\node [scalarnode] at ( 2, 1.5 ) (theta) { $\theta$ };
\node [scalarnode] at ( 4, 1.5) (z) {z};
\node [scalarnode, fill=black!30] at ( 5.5, 1.5) (w) {w};
\node [scalarnode] at ( 8, 1.5) (beta) {$\beta$};
\node [Vnode] at ( 9.5, 1.5) (sigma) {};
\node [] at ( 10, 1.5) () {$\sigma$};

\node[surround, inner sep = .3cm] (f_N) [fit = (z)(w) ] {};
\node[surround, inner sep = .5cm] (f_M) [fit = (f_N)(theta)] {};
\node[surround, inner sep = .3cm] (f_beta) [fit = (beta)] {};

\node [] at (6, 0.5) (M) {\scriptsize $m \in [1,M]$};
\node [] at (5.5, 1) (N) {\scriptsize $n \in [1,N]$};
\node [] at (8.15, 1) () {\scriptsize $k \in [1,K]$};

\path [arrowline] (alpha_s) to (theta); 
\path [arrowline] (theta) to (z); 
\path [arrowline] (z) to (w); 
\path [arrowline] (beta) to (w); 
\path [arrowline] (sigma) to (beta); 

\end{tikzpicture}

%% file: tikz/PGM_NView_LDA.tex
\pgfdeclarelayer{background}
\pgfdeclarelayer{foreground}
\pgfsetlayers{background,main,foreground}

\begin{tikzpicture}

\tikzstyle{surround} = [thick,draw=black,rounded corners=1mm]

\tikzstyle{scalarnode} = [circle, draw, fill=white!11,  
    text width=1.2em, text badly centered, inner sep=2.5pt]

\tikzstyle{scalarnodeCyan} = [circle, draw=cyan, fill=white!11,  
    text width=1.2em, text badly centered, inner sep=2.5pt]
\tikzstyle{discnode}=[rectangle,draw,fill=white!11,minimum size=0.9cm]

\tikzstyle{Vnode}=[circle, radius=1pt,draw,fill=black]
\tikzstyle{vectornode} = [circle, draw, fill=white!11,  
    text width=2.3em, text badly centered, inner sep=2pt]
\tikzstyle{state} = [rectangle, draw, text centered, fill=white, 
    text width=8em, text height=6.7em, rounded corners]

\tikzstyle{arrowline} = [draw,color=black, -latex]
\tikzstyle{carrowline} = [line width=2pt, draw,color=black, -latex]
\tikzstyle{line} = [draw]

\node [Vnode] at ( 0.5, 1.5) (alpha_s){};
\node [] at ( 0, 1.5) (){$\alpha$};
\node [scalarnode] at ( 2, 1.5 ) (theta) { $\theta$ };
\node [scalarnode] at ( 4, 1.5) (z) {z};
\node [scalarnode, fill=black!30] at ( 5.5, 1.5) (w) {w};
\node [scalarnode] at ( 8, 1.5) (beta) {$\beta$};
\node [Vnode] at ( 9.5, 1.5) (sigma) {};
\node [] at ( 10, 1.5) () {$\sigma$};

\node[surround, inner sep = .3cm] (f_N) [fit = (z)(w) ] {};
\node[surround, inner sep = .7cm] (f_M) [fit = (theta)(f_N)] {};
\node[surround, inner sep = .3cm] (f_beta) [fit = (beta)] {};
\node[surround, inner sep = .3cm] (f_V) [fit = (f_beta)(f_N)] {};

\node [] at (6.2, 0.35) (M) {\scriptsize $m \in [1,M]$};
\node [] at (5.5, 1) (N) {\scriptsize $n \in [1,N]$};
\node [] at (8.15, 1) () {\scriptsize $k \in [1,K]$};
\node [] at (8.3, 0.65) (V) {\scriptsize $v \in [1,V]$};

\path [arrowline] (alpha_s) to (theta); 
\path [arrowline] (theta) to (z); 
\path [arrowline] (z) to (w); 
\path [arrowline] (beta) to (w); 
\path [arrowline] (sigma) to (beta); 

\end{tikzpicture}

%% file: tikz/general_PGM.tex
\pgfdeclarelayer{background}
\pgfdeclarelayer{foreground}
\pgfsetlayers{background,main,foreground}

\begin{tikzpicture}

\tikzstyle{surround} = [thick,draw=black,rounded corners=1mm]

\tikzstyle{scalarnode} = [circle, draw, fill=white!11,  
    text width=1.2em, text badly centered, inner sep=2.5pt]

\tikzstyle{scalarnodeCyan} = [circle, draw=cyan, fill=white!11,  
    text width=1.2em, text badly centered, inner sep=2.5pt]
\tikzstyle{discnode}=[rectangle,draw,fill=white!11,minimum size=0.9cm]

\tikzstyle{Vnode}=[circle, radius=1pt,draw,fill=black]
\tikzstyle{vectornode} = [circle, draw, fill=white!11,  
    text width=2.3em, text badly centered, inner sep=2pt]
\tikzstyle{state} = [rectangle, draw, text centered, fill=white, 
    text width=8em, text height=6.7em, rounded corners]

\tikzstyle{arrowline} = [draw,color=black, -latex]
\tikzstyle{carrowline} = [line width=2pt, draw,color=black, -latex]
\tikzstyle{line} = [draw]

\node [Vnode] at ( 0.5, 0) (alpha_s){};
\node [] at ( 0, 0) (){$\alpha_s$};
\node [scalarnode] at ( 2, 0 ) (theta) { $\theta$ };
\node [Vnode] at ( 0.5, 3) (alpha_p1){};
\node [] at ( 0, 3) (){$\alpha_{p}$};
\node [scalarnode] at ( 2 , 3 ) (kappa) { $\kappa$ };
\node [scalarnode] at ( 4, 1.5) (z) {z};
\node [scalarnode, fill=black!30] at ( 5.5, 1.5) (w) {w};

\node [scalarnode] at ( 4, 3) (rho) {$\rho$};
\node [Vnode] at ( 4, 5) (iota1) {};
\node [] at ( 4.5, 5) () {$\iota$};

\node [scalarnode] at ( 8, 1.5+2) (zeta) {$\zeta$};
\node [Vnode] at ( 9.5, 1.5+2) (sigma_p1) {};
\node [] at ( 10, 1.5+2) () {$\sigma_{p}$};

\node [scalarnode] at ( 8, 1.5) (beta) {$\beta$};
\node [Vnode] at ( 9.5, 1.5) (sigma_s1) {};
\node [] at ( 10, 1.5) () {$\sigma_{s}$};

\node[surround, inner sep = .3cm] (f_N) [fit = (z)(w) ] {};
\node[surround, inner sep = .5cm] (f_M) [fit = (f_N)(theta)(rho) ] {};

\node[surround, inner sep = .3cm] (f_zeta) [fit = (zeta) ] {};
\node[surround, inner sep = .3cm] (f_beta) [fit = (beta) ] {};

\node[surround, inner sep = .2cm] (f_V) [fit = (f_N)(rho)(f_zeta)(f_beta)(kappa) ] {};
\node [] at (6, -0.5) (M) {\scriptsize $m \in [1,M]$};
\node [] at (2, 4.1) (V) {\scriptsize $d \in [1,D]$};
\node [] at (5.5, 1) (N) {\scriptsize $n \in [1,N]$};
\node [] at (8.15, 3) () {\scriptsize $t \in [1,T]$};
\node [] at (8.15, 1) () {\scriptsize $k \in [1,K]$};

\path [arrowline] (alpha_s) to (theta); 
\path [arrowline] (alpha_p1) to (kappa); 
\path [arrowline] (theta) to (z); 
\path [arrowline] (kappa) to (z); 
\path [arrowline] (rho) to (z); 
\path [arrowline] (z) to (w); 
\path [arrowline] (iota1) to (rho); 

\path [arrowline] (zeta) to (w); 
\path [arrowline] (sigma_p1) to (zeta); 
\path [arrowline] (beta) to (w); 
\path [arrowline] (sigma_s1) to (beta); 
\end{tikzpicture}

%% file: exp.tex
\vspace{-10pt}
\section{Experiments}
\vspace{-6pt}
\paragraph{Dataset} A dataset of 174 real-world patient discomfort drawings was collected from clinical records with diagnostic labels from medical experts.
Since bilateral diagnostic labels indicates the problem shows in both sides, we preprocess the data breaking all bilateral labels into left side and right side labels.
About $30\%$ of these diagnostic labels appear only once in the dataset.
\vspace{-3pt}
\subsection{Diagnostic Prediction Evaluation} 
\vspace{-3pt}
We randomly split the dataset into two halves.
One half is used for training 
while the other half is used for testing.
We cluster all  painted point locations on the drawing using K-means clustering with $256$ clusters. 
Subsequently, each discomfort drawing is represented using a bag-of-location words.  


The dataset is randomly split 10 times for evaluation and the performance is reported in Table \ref{tab:PredictionPrf} with mean and standard deviation for these 10 groups of experiments with different numbers of shared topics\footnote{For each experiment setting, 10 random seeds were considered and the best result is used.}. The number of private topics is set to $T=5, S=5$ in all the experiments. The hyper-parameters are set to $\alpha_{*}=0.6$, $\sigma_{*}=0.6$ and $\iota_{*}=(5,5)$.
We use the average F-measure on the predicted diagnostic terms to evaluate the prediction performance. The number of predicted diagnostic terms is  determined by mean shift clustering for each test drawing. Additionally, we set a maximum number of 30 labels for both symptom diagnoses and reason diagnoses. 

\begin{table}[h]
\centering
\begin{tabular}{ | c| c | c | c|c | c|}
\hline
& $K=10$ & $K=15$ & $K=20$& $K=30$& $K=50$ \\
\hline
\small Sym F-measure& \small $27.79 \pm 1.41\%$ &\small $28.88 \pm 1.702\%$ &\small $29.95 \pm 1.26\%$ &\small $32.13 \pm 2.18\%$ &\small $32.34 \pm 1.58\%$\\
\hline
\small Rsn F-measure& \small $41.78 \pm 2.09\%$ &\small $42.53 \pm 1.98\%$ &\small $42.82 \pm 1.94\%$ &\small $43.25 \pm 1.59\%$ &\small $43.37 \pm 1.95\%$\\
\hline
\end{tabular}
\caption{\footnotesize Prediction performance for symptom diagnostic labels (Sym) and reason diagnostic labels (Rsn).}
\label{tab:PredictionPrf}
\end{table}
\vspace{-8pt}

\vspace{-5pt}
\begin{table}[h]
\centering
\begin{tabular}{ | c| p{9.9cm} |}
\hline
\multirow{4}{*}{\includegraphics[height=3cm]{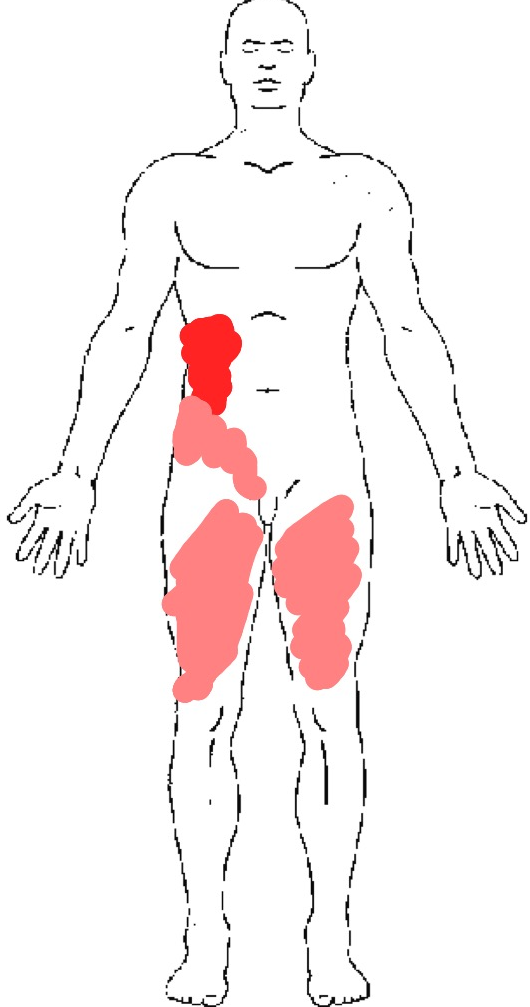}
\includegraphics[height=3cm]{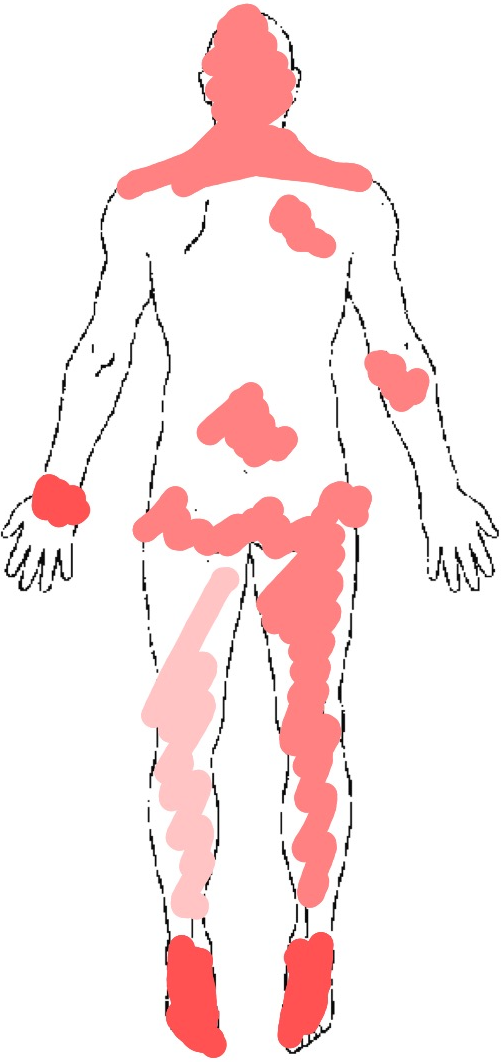}
}
&\scriptsize \textbf{Sym Prd:} {\color{red} R calf dcf,	R front leg dcf,	L  calf dcf, } {\color{blue}  Lumbago, }	{\color{red}L front leg dcf,	} {\color{blue}L back leg dcf, }{\color{red}	R side thigh dcf,	L foot arch dcf,	neck dcf,	L side thigh dcf, }	 {\color{blue} R back leg dcf,}	{\color{red} L hip joint dcf,	L achilles dcf, 	R achilles dcf, 	R upper trapezius dcf, 	R foot arch dcf,	L upper trapezius dcf, R ledgdcf.}\\
&\scriptsize  \textbf{Sym GT:} {\color{red}IBS,} {\color{blue}L back leg dcf, R back leg dcf, } {\color{red} Headache,} {\color{blue} Lumbago,} {\color{red} R Medial elbow dcf; Thoracic spine  dcf; Coccydynia}.\\

& \scriptsize  \textbf{Rsn Prd:} {\color{blue}DLI L5-S1,	R S1 Rdc,}	{\color{red}L S1 Rdc,}	{\color{blue}R L4 Rdc, DLI L4-L5,}	{\color{red}L L5 Rdc,}	{\color{blue}DLI C3-C4,	R C7 Rdc,	 } {\color{red}R L5 Rdc,	DLI L3-L4, } {\color{blue}	L L4 Rdc,	L C7 Rdc,}	{\color{blue}DLI C6-C7,}	{\color{blue}L C4 Rdc,	R C4 Rdc,}	{\color{red}R C6 Rdc,	L C6 Rdc,	DLI C5-C6. }\\

&\scriptsize  \textbf{Rsn GT:}{\color{blue} L C4 Rdc, R C4 Rdc, L C7 Rdc, R C7 Rdc,} {\color{red} R T10 Rdc, }{\color{blue}L L4 Rdc, R L4 Rdc, R S1 Rdc, }{\color{red}Thoracic dysfunction, }{\color{blue}DLI  L4-L5, DLI  L5-S1, DLI  C3-C4,} {\color{red}DLI  T9-T10,} {\color{blue}DLI  C6-C7,}{\color{red} Coxysskada.}\\
\hline
\end{tabular}
\caption{\footnotesize Prediction examples using an unseen discomfort drawing: The left column shows the input discomfort drawing. The right column shows the predicted diagnostic labels using IBTM after \textbf{Prd:} and the ground truth diagnostic labels given by medical experts after \textbf{GT:}. \textbf{Sym} correspond to symptom descriptions which is from the second view in the model and \textbf{Rsn}  correspond to the reason descriptions which is from the third view. Correctly predicted labels are marked in blue, while the wrong ones are marked in red. Rdc stands for Radiculopathy and bcf stands for discomfort in the table.}
\label{tab:PredictionExp}
\end{table}
\vspace{-10pt}

Table \ref{tab:PredictionExp} shows typical examples of test results. 
We found that reasonable diagnostic labels can be suggested using IBTM although the F-measure is not very high. For example, every symptom label that the model generated has the corresponding sign in the drawing. Medical personnels commonly focus more on the pathological diagnoses and do not write all symptom descriptions down especially when the symptoms are typical ones under certain pathological diagnose.  For the reason diagnostic predications, it is challenging but the predicted labels are rather reasonable, even for the predictions that are not matching the ground truth. For example, C6 and C7 nerves have over lapping regions on the shoulder and L5 nerve is between L4 and S4 nerves. Hence, the predicted labels are strong candidates for the given symptoms. 
In the end, manually judging the predicted labels, $80\%$ of the labels are in fact reasonable. The measurement in Table \ref{tab:PredictionPrf} is a rather rough measure without considering more fitting metrics. With a systematic  evaluation standard, for example, considering  the predicted label that has a correspondence on the drawing  as a correct prediction, the F-Measure can be easily recomputed around $70\%$. 
Hence, we believe that with more data and more systematic diagnostic labels, machine learning algorithms can achieve high quality diagnostic predictions. We identify this as an important direction of future work. 


%% file: discussion.tex
\vspace{-5pt}
\section{Discussion}
\vspace{-4pt}
In this paper, we used IBTM for automated assessment of discomfort drawings. A dataset containing real-world discomfort drawings and corresponding diagnostic labels was collected. Reasonable diagnostic predictions were found in the experiments. 
This  preliminary work on this application area shows a promising research direction. We will continue to enlarge and refine the dataset, improve the model, and explore new tools, such as SNOMED CT \footnote{http://www.ihtsdo.org/snomed-ct} for data pre-processing. 
At the same time, we will investigate how to present machine learning results to real-life health care personnel. 
We believe that  applying machine learning for diagnostic prediction on  discomfort drawings may have a significant impact on the health care system. It may lead to decision support systems that can help health care personnel to increase effectiveness and precision in diagnosis and treatment of patients.